\documentclass[runningheads]{llncs}
\usepackage[T1]{fontenc}
\usepackage{booktabs}
\usepackage{amssymb}
\usepackage{multirow}

\usepackage{todonotes}

\newcommand{\bs}{$\blacksquare$}
\newcommand{\rs}{$\square$}
\newcommand{\ct}{\texttt{CheckThat!}}

\usepackage{url}
\usepackage{soul}
\usepackage{xcolor} 
\usepackage{todonotes}
\definecolor{blue}{rgb}{0,0, 0.6}
\definecolor{dkgreen}{rgb}{0,0.6,0}
\definecolor{gray}{rgb}{0.5,0.5,0.5}
\definecolor{mauve}{rgb}{0.58,0,0.82}
\definecolor{mauve}{rgb}{0,0,0}
\definecolor{black}{rgb}{0,0,0}
\definecolor{tri}{rgb}{.25,.88,.82}
\definecolor{lilac}{rgb}{0.85,0.64,0.85}
\definecolor{lightblue}{rgb}{0.53, 0.81, 0.98}
\definecolor{lightskyblue}{rgb}{0.53, 0.81, 0.98}

\begin{document}
\title{The CLEF-2025 CheckThat! Lab: Subjectivity, Fact-Checking,  Claim 
Normalization, and Retrieval}

%
\titlerunning{The CLEF-2025 CheckThat! Lab}
%


\author{%
Firoj Alam\inst{1} \and
Julia Maria Stru\ss\inst{2} \and
Tanmoy	Chakraborty\inst{6}	 \and
Stefan Dietze\inst{7,8} \and
Salim Hafid\inst{12} \and
Katerina Korre\inst{4} \and
Arianna Muti\inst{4} \and
Preslav Nakov\inst{5} \and
Federico Ruggeri\inst{3} \and
Sebastian Schellhammer\inst{7} \and
Vinay Setty\inst{9} \and
Megha Sundriyal\inst{11} \and
Konstantin Todorov\inst{12} \and
Venktesh V \inst{10}
}
\authorrunning{F. Alam et al.}
%
\institute{%
Qatar Computing Research Institute, HBKU, Doha, Qatar \and
University of Applied Sciences Potsdam, Potsdam, Germany \and
DISI, University of Bologna, Bologna, Italy \and
DIT, University of Bologna, Forl\'i, Italy \and
Mohamed bin Zayed University of Artificial Intelligence \and
Indian Institute of Technology Delhi, New Delhi, India \and
GESIS - Leibniz Institute for the Social Sciences, Cologne, Germany \and 
Heinrich-Heine-University D\"usseldorf, Germany \and
Universtity of Stavanger, Norway \and
Delft University of Technology, The Netherlands\and
Indraprastha Institute of Information Technology, New Delhi, India\and University of Montpellier, LIRMM, CNRS, Montpellier, France
\\
\email{https://checkthat.gitlab.io}
}
\vspace{-20mm}
\maketitle              
\begin{abstract}
The \ct~lab aims to advance the development of innovative technologies designed to identify and to counteract online disinformation and manipulation efforts across various languages and platforms. The first five editions of the \ct{} lab focused on the main tasks of the information verification pipeline: check-worthiness, evidence retrieval and pairing, and verification. Since the 2023 edition, the lab has broadened the focus and addressed new problems on auxiliary tasks supporting research and decision-making during the verification process. In the 2025 edition of the lab, we consider tasks at the core of the verification pipeline again as well as auxiliary tasks:
Task~1 is on identification of subjectivity (a follow up of the \ct{} 2024 edition),
Task~2 is on claim normalization, 
Task~3 addresses fact-checking numerical claims, and
Task~4 focuses on scientific web discourse processing.
These tasks represent challenging classification and retrieval problems at the document and at the span level, including multilingual settings. 
\end{abstract}

\section{Introduction}
\label{sec:intro}
\begin{figure}[t]
\centering
\includegraphics[width=\textwidth]{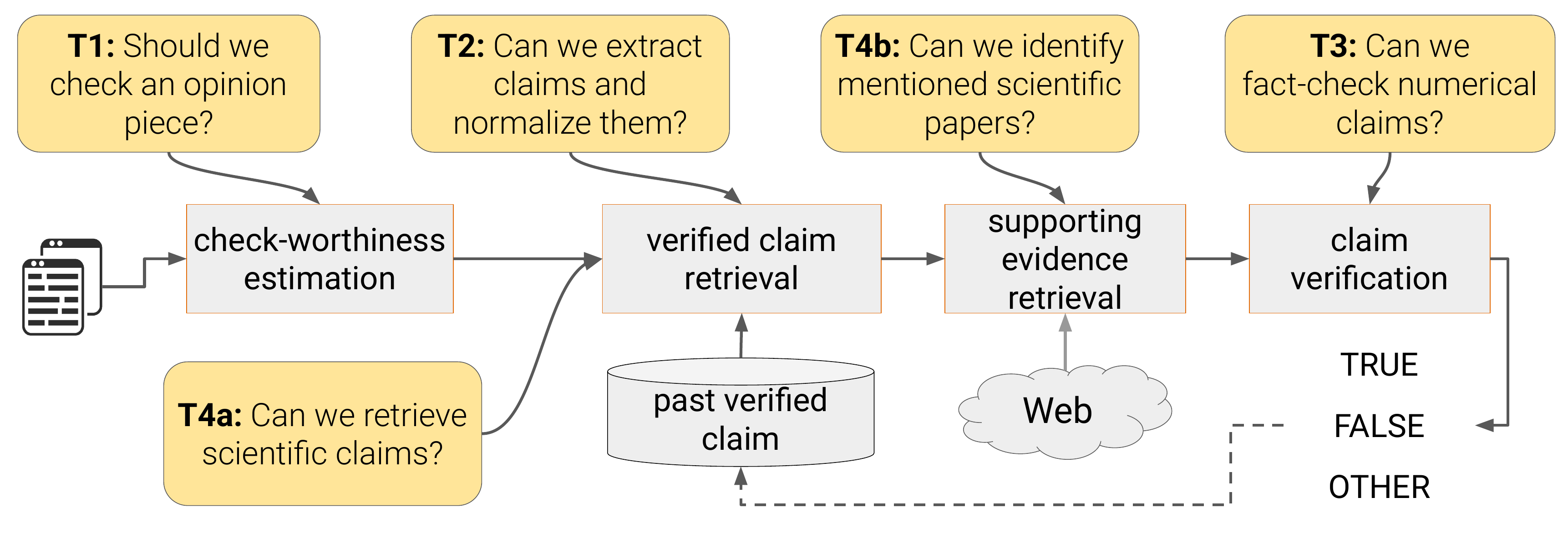}
\caption{Overview of the \ct{} verification pipeline featuring the core tasks along with the 2025 tasks.}
\label{fig:pipeline}
\end{figure}

During its previous seven editions 
(e.g., see \cite{clef-checkthat:2023,CheckThat:ECIR2024}),
the \ct{} lab has fostered the development of technology to assist \textit{fact-checkers} and \textit{journalists} during the main steps of the verification process (see Fig.~\ref{fig:pipeline}). Given a document, or a claim, it first has to be assessed for check-worthiness, i.e.\, whether a journalist should check its veracity - a task 
has been pronounced as being most valued by experts in the past. If this is so, the system needs to retrieve claims verified in the past that could be useful to fact-check the current one. Further evidence to verify the claim could be retrieved from the Web, if necessary. Finally, with the evidence gathered from diverse sources, a decision can be made: whether the claim is factually true or not with various degrees in-between.

The four tasks proposed for the 2025 edition cover three of the four main tasks in the pipeline, highlighted in Figure \ref{fig:pipeline}, either directly or by addressing auxiliary challenges for individual task in more than 20 languages (see in Table~\ref{tab:lang}):



\noindent
\textbf{Task~1 Subjectivity in news articles:} is an auxiliary task related to check-worthiness estimation and asks to spot text that should be processed with specific strategies~\cite{riloff-wiebe-2003-learning}; benefiting the fact-checking pipeline~\cite{DBLP:conf/webmedia/VieiraJCM20,DBLP:journals/information/KasnesisTP21} 
(cf.~Section~\ref{sec:task1}).

\noindent
\textbf{Task~2 Claim Normalization:} is related to the verified claim retrieval and verification task and asks to simplify a noisy social media claim into a normalized form (cf.~Section~\ref{sec:task2}).

\noindent
\textbf{Task~3 Fact-Checking Numerical Claims:} is addressing the last task of the verification pipeline focusing on numerical claims (cf.~Section~\ref{sec:task3}).

\noindent
\textbf{Task~4 Scientific Web Discourse Processing:} is related to the second and third task of the verification pipeline and asks to detect different types of references or mentions to scientific work as well as identifying the original study a social media post refers to
 (cf.~Section~\ref{sec:task4}).

\begin{table}
\centering
\caption{Languages targeted in the four tasks of the \ct{} 2025 edition. White squares indicate languages that are only offered for testing without any training data in these languages.}\label{tab:lang}
\resizebox{\textwidth}{!}{
\begin{tabular}{@{}lccccccccccccccccccccc@{}}
\toprule
 & \textbf{ara} & \textbf{bul} & \textbf{ben} & \textbf{ces} & \textbf{deu} & \textbf{ell} & \textbf{eng} & \textbf{spa} & \textbf{fra} & \textbf{hin} & \textbf{msa} & \textbf{ita} & \textbf{kor} & \textbf{mar} & \textbf{nld} & \textbf{pan} & \textbf{por} & \textbf{ron} & \textbf{tam} & \textbf{tel} & \textbf{tha} \\ \midrule
T1 & \bs & \bs &  &  & \bs &  & \bs & \rs & \rs &  &  & \bs &  &  &  &  &  &  &  &  &  \\ 
T2 & \bs &  & \rs & \rs & \bs & \rs & \bs & \bs & \bs & \bs & \bs &  & \rs & \bs & \rs & \bs & \bs & \rs & \bs & \bs & \bs \\ 
T3 & \bs &  &  &  &  &  & \bs & \bs &  &  &  &  &  &  &  &  &  &  &  &  &  \\ 
T4 &  &  &  &  &  &  & \bs &  &  &  &  &  &  &  &  &  &  &  &  &  &  \\ \bottomrule
\end{tabular}}
\end{table}

\section{Task 1: Subjectivity Detection}
\label{sec:task1}


\paragraph{\textbf{Motivation}.}
Verifiable facts can be conveyed through both objective and subjective statements. Objective statements are straightforward and can be directly verified, while subjective statements require further analysis, such as distilling them into an objective form to assess their claims.

\paragraph{\textbf{Task definition}.}
Given a sentence from a news article, determine whether it is subjective or objective. This is a binary classification task and is offered in Arabic, English, Bulgarian, German, and Italian for mono- and multi-lingual settings. Additionally, unseen languages like French and Spanish are considered for zero-shot setting.

\paragraph{\textbf{Data}.}
For training and validation, we provide $2.6k$ sentences in Arabic, $1.3k$ sentences in Bulgarian, $1.7k$ in English, $1.6k$ in German, and $2.8k$ in Italian from last year's iteration~\cite{clef-checkthat:2024:task2}. 
About 300 new sentences are being collected and labeled for each 
language to be used as novel test sets. The dataset for the multi-lingual setting will be compiled from the individual datasets of the aforementioned languages.

\paragraph{\textbf{Evaluation}.}
We use macro-averaged F$_1$-measure as the evaluation metric. 


\section{Task 2: Claim Normalization}
\label{sec:task2}


\paragraph{\textbf{Motivation}}
With an upsurge in social media, users have been exposed to many misleading claims. However, the ubiquitous noise in these posts makes it difficult for manual fact-checkers to quickly identify specific and precise claims that need verification. It is difficult and time-consuming to extract pertinent claims from such posts. Therefore, we aim to bridge this gap by decomposing social media posts into simpler, more comprehensible forms, which are referred to as \textit{normalized claims} \cite{sundriyal2023chaos}.

\paragraph{\textbf{Task definition}}
Given a noisy, unstructured social media post, the task is to simplify it into a concise form. This is a generation task, where participants are expected to generate a simplified version of the post while preserving the core assertion. The task will be offered in 20 languages: English, Arabic, Bengali, Czech, German, Greek, French, Hindi, Korean, Marathi, Indonesian, Dutch, Punjabi, Polish, Portuguese, Romanian, Spanish, Tamil, Telugu, Thai.

\paragraph{\textbf{Data}} 
The posts originate from various social media platforms, such as Twitter, Reddit, Facebook, etc., and are sourced from the Google Fact-check Explorer API\footnote{\url{https://toolbox.google.com/factcheck/apis}} and the Claim Review Schema.\footnote{\url{https://schema.org/ClaimReview}} Each post is paired with a corresponding normalized claim. We provide train, dev and test data for Arabic, German, English, French, Hindi, Marathi, Indonesian, Punjabi, Portuguese, Spanish, Tamil and Thai. While low-resource languages like Bengali, Czech, Greek, Korean, Romanian, and Telugu are considered for zero-shot settings. The data statistics are provided in Table \ref{tab:task2_stats}.

\begin{table*}[]
\centering
\caption{Dataset statistics for Task 2 (Claim normalization).\label{tab:task2_stats}}
\resizebox{\textwidth}{!}{
    \begin{tabular}{lcccccccccc}
    \toprule
    \multicolumn{1}{c}{\multirow{2}{*}{\textbf{Language}}} & \textbf{Arabic} & \textbf{Bengali} & \textbf{Czech} & \textbf{German} & \textbf{Greek} & \textbf{English} & \textbf{French} & \textbf{Hindi} & \textbf{Korean} & \textbf{Marathi} \\
    
    & \textbf{(ara)}    & \textbf{(ben)}      & \textbf{(ces)}   & \textbf{(deu)}    & \textbf{(ell)}   & \textbf{(eng)}     & \textbf{(fra)}    & \textbf{(hin)}    & \textbf{(kor)}    & \textbf{(mar)}      \\
    \midrule
    \textbf{Train}  & 470  & 0  & 0  & 386 & 0 & 11374  & 1174  & 1081 & 0 & 137  \\
    \textbf{Dev}  & 118             & 0                & 0              & 101  & 0 & 1171 & 147  & 50  & 0               & 50               \\
    \textbf{Test}                                          & 100             & 81               & 123            & 100             & 156            & 1285             & 148             & 100            & 274             & 100
    \\ 
    
    \toprule

    \multirow{2}{*}{\textbf{Language}} & \textbf{Indonesian} & \textbf{Dutch} & \textbf{Punjabi} & \textbf{Polish} & \textbf{Portugese} & \textbf{Romanian} & \textbf{Spanish} & \textbf{Tamil} & \textbf{Telugu} & \textbf{Thai} \\
    
    & \textbf{(msa)}        & \textbf{(nld)}   & \textbf{(pan)}      & \textbf{(pol)}    & \textbf{(por)}  & \textbf{(ron)}      & \textbf{(spa)}     & \textbf{(tam)}    & \textbf{(tel)}     & \textbf{(tha)}  \\
    \midrule

    \textbf{Train}                     & 540                 & 0              & 445              & 163             & 1735               & 0                 & 3458             & 102            & 0               & 244           \\
    \textbf{Dev}                       & 137                 & 0              & 50               & 41              & 223                & 0                 & 439              & 50             & 0               & 61            \\
    \textbf{Test}                      & 100                 & 177            & 100              & 100             & 225                & 141               & 439              & 100            & 116             & 100          
    
    \\ 
\bottomrule
\end{tabular}
}
\end{table*}

\paragraph{\textbf{Evaluation}}
For evaluation, we use METEOR score.

\section{Task 3: Fact-Checking Numerical Claims}
\label{sec:task3}


\paragraph{\textbf{Motivation}}

There has been growing interest in developing tools~\cite{setty2024factcheck}, 
methods~\cite{guo2022survey}, 
and benchmarks~\cite{multifc,schlichtkrull2023averitec} 
to enhance the fact-checking process. Automating fact-checking is challenging, as many claims are complex and require sophisticated reasoning for accurate validation, especially those involving numerical data. Numerical claims often appear more credible due to the \emph{Numeric-Truth effect} \cite{sagara2009consumer}, leading to uncritical acceptance. Recent studies show verifying numerical claims is more difficult than non-numerical ones~\cite{venktesh2024quantemp,Aly:2021:NeurIPs}. For example, the social media claim that ``CDC quietly deletes 6,000 COVID vaccine deaths from its website'' exaggerates a clerical correction, causing unnecessary panic. This demonstrates the need for automated verification of such misleading claims. Therefore, we propose a task dedicated to verifying numerical claims.

\paragraph{\textbf{Task definition}}
This task focuses on verifying claims with numerical quantities and temporal expressions. Numerical claims are defined as those requiring validation of explicit or implicit quantitative or temporal details. Participants must classify each claim as \textit{True, False, or Conflicting} based on a short list of evidence. Each claim is accompanied by the top-100 pieces of evidence retrieved using BM25 from our collection. Participants can use the evidence as they see fit, for numerical reasoning. The task is available in English, Spanish, and Arabic. 
\begin{table}[t]
\footnotesize\centering
\caption{Dataset statistics for task 3}
\label{tab:my_label}
\begin{tabular}{lr}
\toprule
\bf Language   & \bf \# of claims    \\\midrule
English &  15,514 \\
Spanish     &  2000                       \\
Arabic     &    1536                   \\
\bottomrule
\end{tabular}
\end{table}
\paragraph{\textbf{Data}}
 The dataset is collected from various fact-checking domains through Google Fact-check Explorer API\footnote{\url{https://toolbox.google.com/factcheck/apis}}, complete with detailed metadata and an evidence corpus sourced from the web. Our pipeline will filter out numerical claims for the task. An overview of dataset statistics is shown in Table \ref{tab:my_label}.  We will use the English dataset released in \cite{venktesh2024quantemp} and we will further release Arabic and Spanish claims we have identified with corresponding evidence collections.
\paragraph{\textbf{Evaluation}}
We will employ macro-averaged F1 and classwise F1 scores for evaluating claim verification.

\section{Task 4: Scientific Web Discourse Processing (SciWeb)}
\label{sec:task4}

\paragraph{\textbf{Motivation}}
Scientific web discourse, e.g., discourse about scientific claims or resources on the social web, has increased substantially throughout the past years and has been studied across a range of disciplines \cite{dunwoody2021science,bruggemann2020post}. 

However, scientific web discourse is usually informal, with examples such as \textit{``covid vaccines just don't work on children''}, and displays fuzzy/incomplete citation habits, such as \textit{``Stanford study shows that vaccines don't work''} where the actual study is never cited through explicit references. This poses challenges both from a computational perspective when mining social media or computing Altmetrics but also from a societal perspective leading to poorly informed online debates \cite{rocha2021impact}. 
Based on this motivation, we introduce two tasks: scientific web discourse detection (\textbf{Subtask 4.1}), and claim-source retrieval (\textbf{Subtask 4.2}).

\paragraph{\textbf{Subtask 4.1: SciWeb Discourse Detection}}
This task aims at classifying the different forms of science-related online discourse as introduced in \cite{hafid2022scitweets}. Namely, given a tweet, this multilabel task aims at detecting if a tweet contains a scientific claim (C1.1) or scientific reference (C1.2) or is referring to science contexts or entities (C1.3). 
This task will utilize the SciTweets corpus \cite{hafid2022scitweets}, a manually annotated gold standard of 1,261 tweets spanning the three distinct categories of science-relatedness for training. An additional test set of 500-1000 tweets following the same process will be released. 
We will use the macro-averaged $F_{1}$ score to evaluate and compare submissions to this task. 

\paragraph{\textbf{Subtask 4.2: SciWeb Claim-Source Retrieval}}
Given a tweet containing a scientific claim (C1.1) and an informal reference to a scientific paper, this task aims at retrieving the scientific paper that serves as the source for the claim from a given a pool of candidate scientific papers. 
For training and testing, a ground truth dataset containing tweet-study pairs will be provided. Using the Altmetric corpus\footnote{\url{https://www.altmetric.com}} and a pretrained classifier \cite{hafid2022scitweets}, we collect tweets that contain both scientific claims and explicit study references, where explicit references such as URIs will be removed from the test set. We collect scientific papers from the CORD-19 corpus \cite{wang-etal-2020-cord}, where each data point contains the title, abstract, and metadata of a paper related to COVID-19 research. Our final collection comprises 15,967 tweet-study pairs, where each tweet points to an academic paper from the CORD-19 corpus.
We will use the Mean Reciprocal Rank score (\textit{MRR@5}) to evaluate submissions to this task.

\section{Related Work}
\paragraph{\textbf{Fact-checking}}
The problem of misleading information spreading over online has been a major problem and there has been significant research effort in the past years including multilingual resource development to support fact-checker, journalist \cite{alam-etal-2021-fighting-covid,clef-checkthat:2022:task3}, evidence retrieval \cite{malviya-katsigiannis-2024-evidence,zheng-etal-2024-evidence,10.1145/3477495.3531827}, automated fact-checking pipeline \cite{guo-etal-2022-survey,ijcai2021p619}. The advancement of the fact-checking focused not only on textual modality but also visual and multimodal inputs \cite{cekinel2024multimodal}, which are significantly benefiting fact-checking and content moderation tools \cite{yang2024factcheckingtoolshelpfulexploration,horta2023automated}. For example, \cite{horta2023automated} investigated 412M facebook comments and  measured the impact of automated content moderation on subsequent
rule-breaking behavior and engagement. They finding suggest that content moderation increases adherence to community guidelines. 

Subjectivity is an important aspect of the entire fact-checking pipeline. Identifying subjectivity allows the system to analyze tone, intent, and context, which are crucial for detecting subtle misinformation or persuasive techniques often employed in propaganda or misleading claims. By considering subjective elements, such as emotional appeals or rhetorical framing, the pipeline can deliver a more comprehensive evaluation of content \cite{vargas-etal-2024-improving,pardawalasubjective,clef-checkthat:2023:task2}.

Claims on social media are often noisy. The noisy nature of social media posts makes it challenging to identify important claims that require manual fact-checking. To streamline this process, research efforts have focused on identifying the core assertions of claims and presenting them in an understandable form \cite{sundriyal-etal-2023-chaos,wuehrl-etal-2023-entity,cheema-etal-2022-mm}. Claims often include temporal references, statistical data, time intervals, and comparable values, which are particularly challenging to identify and analyze. Recent research has made significant progress in addressing these challenges by developing resources sourced from various fact-checking platforms and analysts reports, with a focus on both financial and general domains \cite{shah-etal-2024-numerical,venktesh2024quantemp}. Another area of research has emerged to analyze scientific articles, focusing on identifying key claims and aggregating them. This is a complex task, as scientific ideas are typically conveyed in a more intricate manner than the general text found in newspapers and Wikipedia articles \cite{Wei2023ClaimDistiller,wadden2020fact,wuehrl2024makes}. 

In this iteration of the \ct{} Lab, we explore the relevant research areas discussed above. In the following section, we highlight related shared tasks organized in recent years. 


\paragraph{\textbf{Related Shared Tasks}}
The \ct{} Lab task is closely related to several SemEval tasks, including determining rumor veracity~\cite{derczynski2017semeval,gorrell2019semeval}, stance detection~\cite{mohammad2016semeval}, fact-checking in community question-answering forums~\cite{mihaylova2019semeval}, and propaganda detection in various contexts such as English news articles~\cite{da2020semeval}, English memes~\cite{dimitrov2021semeval2021,dimitrov2024semeval}, and Arabic news paragraphs, social media posts, and memes~\cite{hasanain-etal-2023-araieval,hasanain2024araieval}. Additionally, multilingual persuasion techniques in online news have been explored in~\cite{piskorski2023semeval}.
The tasks in \ct{} Lab is also related to the FEVER tasks~\cite{thorne2018fever} on fact extraction and verification, the Fake News Challenge~\cite{hanselowski-etal-2018-retrospective,FNC1,malliga2023overview}, and the detection of online hostile posts~\cite{patwa2021overview}. Furthermore, it aligns with the FakeNews task at MediaEval~\cite{pogorelov2020fakenews}.
Recent shared tasks include claim detection and span identification in social media posts~\cite{sundriyal2023overview} and multimodal fake news detection~\cite{suryavardan2022findings}, further emphasizing the growing focus on analyzing and verifying information across diverse modalities and platforms.

\section{Conclusions}
\label{sec:conclusions}

The paper presented the setup and data for the four tasks proposed in the 2025 edition, addressing three of the four main stages of the verification pipeline, either directly or by tackling auxiliary challenges specific to individual tasks. Aligned with the CLEF mission, most tasks are designed to support diverse multi- and cross-lingual setups by being offered in multiple languages. This year, the tasks are available in 20 languages. As in previous editions, the evaluation frameworks for all tasks are freely released to the community to encourage the development of effective technologies for combating disinformation and misinformation.

\section*{Acknowledgments}
\begin{footnotesize}
The work of Julia Maria Stru\ss has partially been funded by the BMBF (German Federal Ministry of Education and Research) under grant no. 01FP20031J. 
The work by F. Alam has partially been funded by the NPRP grant 14C-0916-210015 from the Qatar National Research Fund part of Qatar Research Development and Innovation Council (QRDI). The work of Stefan Dietze, Konstantin Todorov, Salim Hafid and Sebastian Schellhammer is partially funded under the AI4Sci grant, co-funded by MESRI (France, grant
UM-211745), BMBF (Germany, grant 01IS21086), and the French National Re-
search Agency (ANR).
The responsibility for the contents of this publication lies with the authors.

\end{footnotesize}
%
%
%
\bibliographystyle{splncs04}
\bibliography{bib/task_5,bib/clef20_checkthat,bib/clef19_checkthat,bib/clef18_checkthat,bib/clef21_checkthat,bib/clef22_checkthat,bib/clef23_checkthat,bib/custom,bib/sigproc,bib/task6, bib/task2,bib/task3, bib/clef24_checkthat}
\end{document}